%% file: cvpr2023-777.tex
\documentclass[10pt,twocolumn,letterpaper]{article}
\input{header/incl}         %

\def\cvprPaperID{777}       %
\def\confName{CVPR}
\def\confYear{2023}

\begin{document}
\input{header/info}         %
\input{sections/0-abs}      %
\input{sections/1-intro}    %
\input{sections/2-related}  %
\input{sections/3-method}   %
\input{sections/4-exper}    %
\input{sections/5-concl}    %

\clearpage
{\small
\bibliographystyle{ieee_fullname}
\bibliography{abbr,ref}
}

\end{document}

%% file: header/incl.tex
\usepackage{cvpr}              %

\usepackage{enumitem}
\usepackage[accsupp]{axessibility}  %

\usepackage{algpseudocode}
\usepackage[table,xcdraw]{xcolor}
\definecolor{commgreen}{HTML}{009933}
\usepackage[linesnumbered,ruled,vlined]{algorithm2e}

\SetCommentSty{licommfont}

\setenumerate[1]{itemsep=0pt,partopsep=0pt,parsep=\parskip,topsep=2.5pt}
\setitemize[1]{itemsep=1pt,partopsep=0pt,parsep=\parskip,topsep=1.5pt}
\setdescription{itemsep=0pt,partopsep=0pt,parsep=\parskip,topsep=2.5pt}
\usepackage{booktabs}
\usepackage{multirow}
\usepackage{mathrsfs}
\usepackage{dsfont}
\usepackage{threeparttable}
\usepackage{graphicx}
\usepackage{amsmath}
\usepackage{amssymb}
\usepackage[pagebackref,breaklinks,colorlinks,urlcolor=blue]{hyperref}
\usepackage{float}
\usepackage{microtype}
\usepackage{mathtools}
\usepackage[geometry]{ifsym}

\usepackage[capitalize]{cleveref}
\Crefname{section}{Section}{Sections}
\crefname{section}{Sec.}{Secs.}
\Crefname{table}{Table}{Tables}
\crefname{table}{Tab.}{Tabs.}

\newcommand\bdtitle[1]{\noindent\textbf{#1}}
\newcommand\topr{\toprule[1.2pt]}
\newcommand\toprr{\toprule[0.85pt]}
\newcommand\midr{\toprule[0.65pt]}
\newcommand\bottomr{\bottomrule[1.2pt]}

\newcolumntype{C}[1]{>{\centering\arraybackslash}p{#1}}
\newcolumntype{L}[1]{>{\raggedleft\arraybackslash}p{#1}}
\newcolumntype{R}[1]{>{\raggedright\arraybackslash}p{#1}}

\newcommand{\ssymbol}[1]{^{\@fnsymbol{#1}}}

\usepackage{rotating}

\usepackage{xcolor}

\definecolor{fg1_blue}{RGB}{68,114,196}
\definecolor{fg1_green}{RGB}{83,130,53}
\definecolor{fg3_yellow}{RGB}{254,255,3}
\definecolor{fg3_blue}{RGB}{152,145,163}
\definecolor{fg3_orange}{RGB}{227,147,124}
\definecolor{fg5_blue}{RGB}{0,84,204}
\definecolor{fg5_red}{RGB}{219,24,24}

%% file: header/info.tex
\title{\vspace{-0.85cm}
\textit{Less is More}: Reducing Task and Model Complexity \\ for 3D Point Cloud Semantic Segmentation
\vspace{-0.55cm}}

\author{Li Li$^1$ \qquad Hubert P. H. Shum$^1$ \qquad Toby P. Breckon$^{1,2}$\\
	Department of \{Computer Science$^{1}$ $|$ Engineering$^{2}$\}, Durham University, UK\\
	{\tt\small \{li.li4,\ hubert.shum,\ toby.breckon\}@durham.ac.uk}}

\maketitle 

\newcommand{\ourmodel}{LiM3D}
\newcommand{\ourmodelsdsc}{LiM3D+SDSC}
\newcommand{\samplfull}{Spatio-Temporal Redundant Frame Downsampling}
\newcommand{\samplshort}{ST-RFD}
\newcommand{\ourupl}{Voxel $\text{U}^2\text{PL}$}
\newcommand{\validset}{\textit{validation} set}
\newcommand{\trainset}{\textit{training} set}
\newcommand{\testset}{\textit{test} set}

%% file: sections/0-abs.tex
\begin{abstract}
    \vspace{-0.2cm}   
    \noindent
    Whilst the availability of 3D LiDAR point cloud data has significantly grown in recent years, annotation remains expensive and time-consuming, leading to a demand for semi-supervised semantic segmentation methods with application domains such as autonomous driving. Existing work very often employs relatively large segmentation backbone networks to improve segmentation accuracy, at the expense of computational costs. In addition, many use uniform sampling to reduce ground truth data requirements for learning needed, often resulting in sub-optimal performance. To address these issues, we propose a new pipeline that employs a smaller architecture, requiring fewer ground-truth annotations to achieve superior segmentation accuracy compared to contemporary approaches. This is facilitated via a novel Sparse Depthwise Separable Convolution module that significantly reduces the network parameter count while retaining overall task performance. To effectively sub-sample our training data, we propose a new {\samplfull} ({\samplshort}) method that leverages knowledge of sensor motion within the environment to extract a more diverse subset of training data frame samples. To leverage the use of limited annotated data samples, we further propose a soft pseudo-label method informed by LiDAR reflectivity. Our method outperforms contemporary semi-supervised work in terms of mIoU, using less labeled data, on the SemanticKITTI (59.5@5\%) and ScribbleKITTI (58.1@5\%) benchmark datasets, based on a 2.3$\times$ reduction in model parameters and 641$\times$ fewer multiply-add operations whilst also demonstrating significant performance improvement on limited training data (i.e., Less is More).

    \vspace{-0.5cm}
\end{abstract}

%% file: sections/1-intro.tex
\vspace{-0.3cm}   
\section{Introduction}
\vspace{-0.1cm}

\noindent
3D semantic segmentation of LiDAR point clouds has played a key role in scene understanding, facilitating applications such as autonomous driving~\cite{hu2020randlanet,hou2022pointtovoxel,jaritz2021xmuda,Unal_2022_CVPR,yi2021complete,zhu2021cylindrical, abarghouei19depth} and robotics~\cite{milioto2019rangenet, wu2018squeezesega, wu2019squeezesegv2a,alonso20203dmininet}. However, many contemporary methods require relatively large backbone architectures with millions of trainable parameters requiring  many hundred gigabytes of annotated data for training at a significant computational cost. Considering the time-consuming and costly nature of 3D LiDAR annotation, such methods have become less feasible for practical deployment.  

\begin{figure}[tp]
    \hspace*{-15pt}
    \centering
    \includegraphics[width=0.53\textwidth]{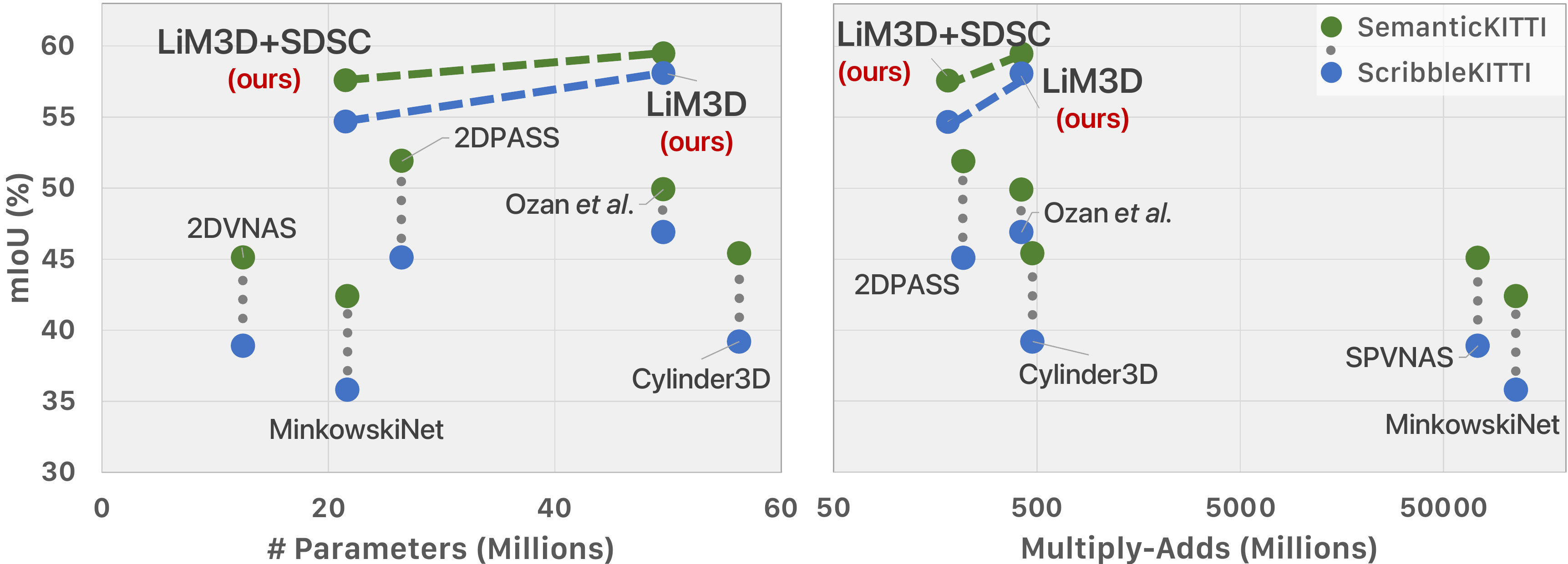}
    \caption{mIoU performance (\%) against parameters and multiply-add operations on SemanticKITTI (fully annotated) and ScribbleKITTI (weakly annotated) under the 5\% sampling protocol.
    \vspace{-0.2cm}
    }
    \label{fig:computaion_cost}
\end{figure}

Existing supervised 3D semantic segmentation methods~\cite{wu2018squeezesega,wu2019squeezesegv2a,choy20194d,milioto2019rangenet,tang2020searching,xu2020squeezesegv3,kochanov2020kprnet,zhu2021cylindrical,yan20222dpass} primarily focus on designing network architectures for densely annotated data. To reduce the need for large-scale data annotation, and inspired by similar work in 2D~\cite{chen2021semisupervised, oord2019representation, wang2022semisupervised}, recent 3D work proposes efficient ways to learn from weak supervision~\cite{Unal_2022_CVPR}. However, such methods still suffer from high training costs and inferior on-task performance. To reduce computational costs, a 2D projection-based point cloud representation is often considered~\cite{alonso20203dmininet, cortinhal2020salsanext, kochanov2020kprnet, milioto2019rangenet, wu2018squeezesega, wu2019squeezesegv2a, xu2020squeezesegv3, zhang2020polarneta}, but again at the expense of significantly reduced on-task performance. As such, we observe a gap in the research literature for the design of semi or weakly supervised methodologies that employ a smaller-scale architectural backbone, hence facilitating improved training efficiency whilst also reducing their associated data annotation requirements.

In this paper, we propose a semi-supervised methodology for 3D LiDAR point cloud semantic segmentation. Facilitated by three novel design aspects, our \textit{Less is More} (LiM) based methodologies require \textit{less} training data and \textit{less} training computation whilst offering (\textit{more}) improved accuracy over contemporary state-of-the-art approaches (see~\cref{fig:computaion_cost}).

Firstly, from an architectural perspective, we propose a novel \textbf{Sparse Depthwise Separable Convolution (SDSC)} module, which substitutes traditional sparse 3D convolution into existing 3D semantic segmentation architectures, resulting in a significant reduction in trainable parameters and numerical computation whilst maintaining on-task performance (see ~\cref{fig:computaion_cost}). Depthwise Separable Convolution has shown to be very effective within image classification tasks~\cite{chollet2017xception}. Here, we tailor a sparse variant of 3D Depthwise Separable Convolution for 3D sparse data by first applying a single submanifold sparse convolutional filter~\cite{graham2017submanifold,graham20183d} to each input channel with a subsequent pointwise convolution to create a linear combination of the sparse depthwise convolution outputs. This work is the first to attempt to introduce depthwise convolution into the 3D point cloud segmentation field as a conduit to reduce model size. Our SDSC module facilitates a 50\% reduction in trainable network parameters without any loss in segmentation performance.

Secondly, from a training data perspective, we propose a novel \textbf{{\samplfull} ({\samplshort})} strategy that more effectively sub-samples a set of diverse frames from a continuously captured LiDAR sequence in order to maximize diversity within a minimal training set size. We observe that continuously captured LiDAR sequences often contain significant temporal redundancy, similar to that found in video ~\cite{akramullah2014digital}, whereby temporally adjacent frames provide poor data variation. On this basis, we propose to compute the temporal correlation between adjacent frame pairs, and use this to select the most informative sub-set of LiDAR frames from a given sequence. Unlike passive sampling (\eg, uniform or random sampling), our active sampling approach samples frames from each sequence such that redundancy is minimized and hence training set diversity is maximal. When compared to commonplace passive random sampling approaches~\cite{jiang2021guided, kong2022lasermix, Unal_2022_CVPR}, {\samplshort} explicitly focuses on extracting a diverse set of training frames that will hence maximize model generalization.

Finally, in order to employ semi-supervised learning, we propose a soft pseudo-label method informed by the LiDAR reflectivity response, thus maximizing the use of any annotated data samples. Whilst directly using unreliable soft pseudo-labels generally results in performance deterioration~\cite{arazo2020pseudolabeling}, the voxels corresponding to the unreliable predictions can instead be effectively leveraged as negative samples of unlikely categories.  Therefore, we use cross-entropy to separate all voxels into two groups, \ie, a reliable and an unreliable group with low and high-entropy voxels respectively. We utilize predictions from the reliable group to derive positive pseudo-labels, while the remaining voxels from the unreliable group are pushed into a FIFO category-wise memory bank of negative samples~\cite{alonso2021semisuperviseda}. To further assist semantic segmentation of varying materials in the situation where we have weak/unreliable/no labels, we append the reflectivity response features onto the point cloud features, which again improve segmentation results.

We evaluate our method on the SemanticKITTI~\cite{behley2019semantickittia} and ScribbleKITTI~\cite{Unal_2022_CVPR} {\validset}. Our method outperforms contemporary state-of-the-art semi-~\cite{kong2022lasermix,jiang2021guided} and weakly-~\cite{Unal_2022_CVPR} supervised methods and offers \textit{more} in terms of performance on limited training data, whilst using \textit{less} trainable parameters and \textit{less} numerical operations (\textit{Less is More}). %

Overall, our contributions can be summarized as follows:
\begin{itemize}
    \item A novel methodology for semi-supervised 3D LiDAR semantic segmentation that uses significantly \textit{less} parameters and offers (\textit{more}) superior accuracy.\footnote{{Full source code: \url{https://github.com/l1997i/lim3d/}.}}
    
    \item A novel Sparse Depthwise Separable Convolution (SDSC) module, to reduce trainable network parameters, and to both reduce the likelihood of over-fitting and facilitate a deeper network architecture.
    
    \item A novel {\samplfull} ({\samplshort}) strategy, to extract a maximally diverse data subset for training by removing temporal redundancy and hence future annotation requirements.
    
    \item A novel soft pseudo-labeling method informed by LiDAR reflectivity as a proxy to in-scene object material properties,  facilitating effective use of limited data annotation. %
\end{itemize}

%% file: sections/2-related.tex
\section{Related Work}

\bdtitle{Semi-supervised learning (SSL) LiDAR semantic segmentation} is a special instance of weak supervision that combines a small amount of labeled, with a large amount of unlabeled point cloud during training. Numerous approaches have been explored for LiDAR semantic segmentation. %
Projection-based approaches~\cite{kong2022lasermix,wu2018squeezesega,wu2019squeezesegv2a,milioto2019rangenet,kochanov2020kprnet,xu2020squeezesegv3,liong2020amvnet} make full use of 2D-convolution kernels by using range or other 2D image-based spherical coordinate representations of point clouds. Conversely, voxel-based approaches~\cite{tang2020searching,zhu2021cylindrical,kong2022lasermix,Unal_2022_CVPR} transform irregular point clouds to regular 3D grids and then apply 3D convolutional neural networks with a better balance of the efficiency and effectiveness. Pseudo-labeling is generally applied to alleviate the side effect of intra-class negative pairs in feature learning from the teacher network~\cite{jiang2021guided,yan2021sparse,Unal_2022_CVPR,kong2022lasermix}. However, such methods only utilize samples with reliable predictions and thus ignore the valuable information that unreliable predictions carry. In our work, we combined a novel SSL framework with the mean teacher paradigm ~\cite{tarvainen2017mean}, demonstrating the utilization of unreliable pseudo-labels to improve segmentation performance.

\begin{figure*}[thp]
    \centering
    \includegraphics[width=1\textwidth]{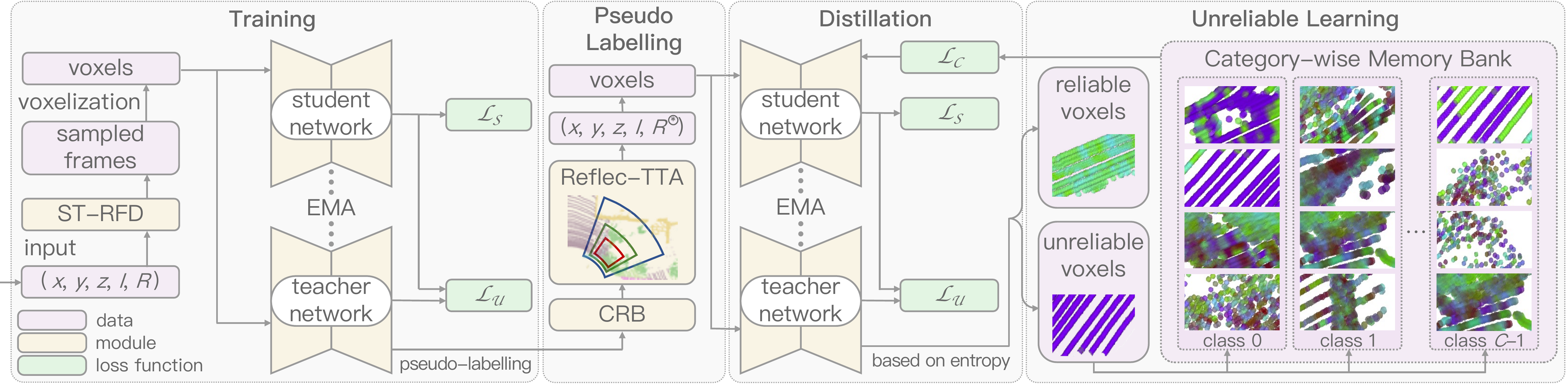}
     \caption{Our proposed architecture for unreliable pseudo-labels LiDAR semantic segmentation involves three stages: training, pseudo-labeling, and distillation with unreliable learning. We apply {\samplshort} sampling before training the Mean Teacher on available annotations.}
    \vspace{-0.3cm}
    \label{fig:model}
\end{figure*}

\bdtitle{Depthwise separable convolution}\cite{sifre2014rigidmotiona} is a depthwise convolution followed by a pointwise convolution, to reduce both model size and complexity. Being a more computationally efficient alternative than standard convolution, it is used for mobile applications~\cite{howard2017mobilenets,sandler2018mobilenetv2,howard2019searching} and %
hardware accelerators~\cite{masters2021making}. Furthermore, it is a building block of Xception~\cite{chollet2017xception}, a deep convolutional neural network architecture that achieves state-of-the-art performance on the ImageNet~\cite{deng2009imagenet} classification task, via more efficient use of model parameterization. In this work, we propose a novel sparse variant of depthwise separable convolution, which has both the efficiency advantages of depthwise separable convolution and those of sparse convolution for processing spatially-sparse data~\cite{graham20183d}.

\bdtitle{Temporal redundancy} is highly prevalent within video ~\cite{zhu2017deep,wang2022longshort} and radar~\cite{li2022exploiting} sequences alike. Existing semi-supervised 3D LiDAR segmentation methods~\cite{kong2022lasermix,Unal_2022_CVPR} utilize a passive uniform sampling strategy to filter unlabeled points from a fully-labeled point cloud dataset. Active learning frameworks handle the redundancy to reduce annotation or training efforts by selecting informative and diverse sub-scenes for label acquisition\cite{duong2017reducing,hu2022lidal,wu2021redala}.
We propose a novel temporal-redundancy-based sampling strategy with comparable time cost to uniform sampling, 
to reduce the inter-frame spatio-temporal redundancy and maximize data diversity.%

%% file: sections/3-method.tex
\vspace{-0.2cm}   
\section{Methodology}
\vspace{-0.2cm}

\noindent
We first present an overview of the mean teacher framework we employ (\cref{sec:mean-teacher}) and then explain our use of unreliable pseudo-labels informed by LiDAR reflectivity for semi-supervised learning (\cref{sec:learning-from-unreliable-pseudo-labels}). Subsequently, we detail our {\samplshort} strategy for dataset diversity (\cref{sec:MRFD}) and finally our parameter-reducing SDSC module (\cref{sec:sdsc}).

Formally, given a LiDAR point cloud $P=\{\textbf{p} \mid \textbf{p}=(x,y,z,I,R) \in \mathbb{R}^5\}$ where $(x,y,z)$ is a 3D coordinate, $I$ is intensity and $R$ is reflectivity, our goal is to train a semantic segmentation model by leveraging both a large amount of unlabeled $U = \{\textbf{p}_i^u\}_{i=1}^{N_u} \varsubsetneq P$ and a smaller set of labeled data $V = \{(\textbf{p}_i^v, \textbf{y}_i^v)\}_{i=1}^{N_v} \varsubsetneq P$.

Our overall architecture involves three stages (\cref{fig:model}): (1) \textbf{Training:} we utilize reflectivity-prior descriptors and adapt the Mean Teacher framework to generate high-quality pseudo-labels; (2) \textbf{Pseudo-labeling:} we fix the trained teacher model prediction in a class-range-balanced~\cite{Unal_2022_CVPR} manner, expanding dataset with Reflectivity-based Test Time Augmentation (Reflec-TTA) during test time; (3) \textbf{Distillation with unreliable predictions:} we train on the generated pseudo-labels, and utilize unreliable pseudo-labels in a category-wise memory bank for improved discrimination.

\subsection{Mean Teacher Framework}
\label{sec:mean-teacher}

We introduce weak supervision using the Mean Teacher framework~\cite{tarvainen2017mean}, which %
avoids the prominent slow training issues associated with Temporal Ensembling~\cite{laine2017temporal}. This framework consists of two models of the same architecture known as the student and teacher respectively, for which we utilize a Cylinder3D~\cite{zhu2021cylindrical}-based segmentation head $f$. The weights of the student model $\theta$ are updated via standard backpropagation, while the weights of the teacher model $\theta^*$ are updated by the student model through Exponential Moving Averaging (EMA):
\begin{equation}\label{equ:ema}
    \theta_{t+1}^* = \kappa  \theta_t^* + (1-\kappa ) \theta_{t+1}, \quad t \in \{0,1,\cdots T-1\},
\end{equation}
where $\kappa$ denotes a smoothing coefficient to determine update speed, and $T$ is the maximum time step.

During training, we train a set of weakly-labeled point cloud frames with voxel-wise inputs generated via asymmetrical 3D convolution networks~\cite{zhu2021cylindrical}. For every point cloud, our optimization target is to minimize the overall loss:
\begin{equation}\label{equ:overall_loss}
    \mathcal{L} = \mathcal{L}_S + \lambda_U \mathcal{L}_U + g \lambda_C \mathcal{L}_C,
\end{equation}
where $\mathcal{L}_S$ and $\mathcal{L}_U$ denote the losses applied to the supervised and unsupervised set of points respectively, $\mathcal{L}_C$ denotes the contrastive loss to make full use of unreliable pseudo-labels, $\lambda_U$ is the weight coefficient of $\mathcal{L}_U$ to balance the losses, and $g$ is the gated coefficient of $\mathcal{L}_C$. $g$ equals $\lambda_C$ if and only if it is in the distillation stage.
We use the consistency loss (implemented as a Kullback-Leibler divergence loss~\cite{hou2022pointtovoxel}), lovasz softmax loss~\cite{berman2018lovaszsoftmax}, and the voxel-level InfoNCE~\cite{oord2019representation} as $\mathcal{L}_U$, $\mathcal{L}_S$ and $\mathcal{L}_C$ respectively.

We first generate our pseudo-labels for the unlabeled points via the teacher model. Subsequently, we generate reliable pseudo-labels in a class-range-balanced (CRB)~\cite{Unal_2022_CVPR} manner, and utilize the qualified unreliable pseudo-labels as negative samples in the distillation stage. Finally, we train the model with both reliable and qualified unreliable pseudo-labels to maximize the quality of the pseudo-labels.

\subsection{Learning from Unreliable Pseudo-Labels}
\label{sec:learning-from-unreliable-pseudo-labels}

\begin{figure}[thp]
    \centering
    \includegraphics[width=0.48\textwidth]{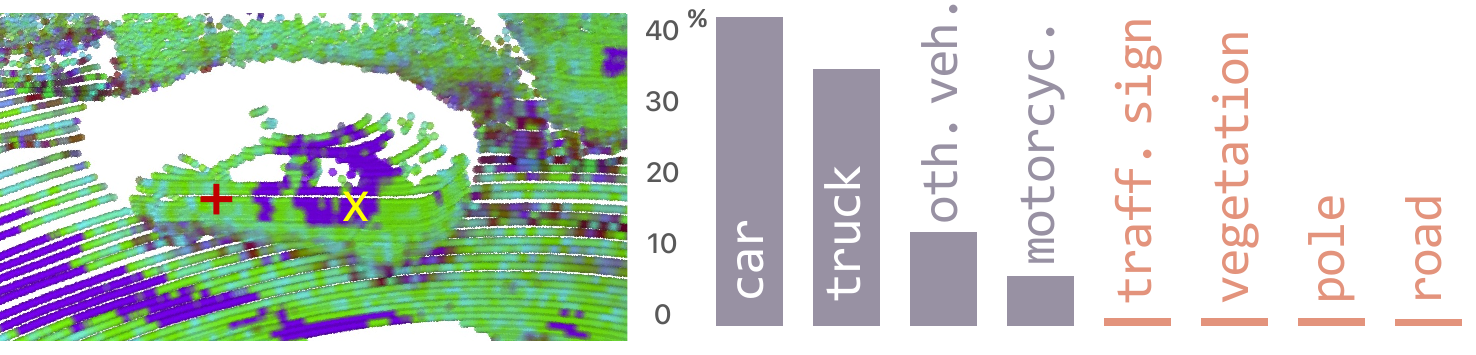}
    \caption{Illustration on unreliable pseudo-labels. Left: entropy predicted from an unlabeled point cloud, with lower entropy corresponding to greener color. Right: Category-wise probability of an unreliable prediction \colorbox[RGB]{115,0,231}{\textcolor{fg3_yellow}{\textsf{X}}}, only \textcolor{fg3_blue}{top-4} and \textcolor{fg3_orange}{last-4} probabilities shown.}
    \label{fig:unreliable}
    \vspace{-8pt}
\end{figure}

\noindent
Unreliable pseudo-labels are frequently eliminated from semi-supervised learning tasks or have their weights decreased to minimize performance loss~\cite{sajjadi2016regularization,xie2020selftraining,zou2021pseudoseg,jiang2021guided,yang2022stb,Unal_2022_CVPR}. In line with this idea, we utilize CRB method~\cite{Unal_2022_CVPR} to first mask off unreliable pseudo-labels and then subsequently generate high-quality reliable pseudo-labels. 

However, such a simplistic discarding of unreliable pseudo-labels may lead to valuable information loss as it is clear that unreliable pseudo-labels (\ie, the corresponding voxels with high entropy) can offer information that helps in discrimination. Voxels that correlate to unreliable predictions can alternatively be thought as negative samples for improbable categories~\cite{wang2022semisupervised}, although performance would suffer if such unreliable predictions are used as pseudo-labels directly~\cite{arazo2020pseudolabeling}. 
As shown in~\cref{fig:unreliable}, the unreliable pseudo predictions show a similar level of confidence on \texttt{car} and \texttt{truck} classes, whilst being sure the voxel cannot be \texttt{pole} or \texttt{road}. 
Thus, together with the use of CRB for high-quality reliable pseudo-labels, we also ideally want to make full use of these remaining unreliable pseudo-labels rather than simply discarding them. Following~\cite{wang2022semisupervised}, we propose a method to leverage such unreliable pseudo-labels for 3D voxels as negative samples. However, to maintain a stable amount of negative samples, we utilize a category-wise memory bank $\mathcal{Q}_c$ (FIFO queue, \cite{wu2018unsupervised}) to store all the negative samples for a given class $c$. As negative candidates in some specific categories are severely limited in a mini-batch due to the long-tailed class distribution of many tasks (\eg autonomous driving), without such an approach in place we may instead see the gradual dominance of large and simple-to-learn classes within our generated pseudo-labels.

Following ~\cite{oord2019representation,he2020momentum}, our method has three prerequisites, \ie, anchor voxels, positive candidates, and negative candidates. They are obtained by sampling from a particular subset, constructed via~\cref{eq:anchor_voxels_sampling_l} and ~\cref{equ:postive_samples}, in order to reduce overall computation. In particular, the set of features of all candidate anchor voxels for class $c$ is denoted as:
\begin{equation} \label{eq:anchor_voxels_sampling_l}
    \mathcal{A}_c=\left\{\mathbf{E}_{a,b} \mid y^{*}_{a,b}=c, p_{a,b}(c)>\delta_p\right\},
\end{equation}
where $\mathbf{E}_{a,b}$ is the feature embedding for the $a$-th point cloud frame at voxel $b$, $\delta_p$ is the positive threshold of all classes, $p_{a,b}(c)$ is the softmax probability by the segmentation head at $c$-th dimension. 
$y^{*}_{a,b}$ is set to the ground truth label $y^{*}_{a,b}$ if the ground truth is available, otherwise, $y^{*}_{a,b}$ is set to the pseudo label $\hat{y}_{a,b}$, due to the absence of ground truth.

The positive sample is the common embedding center of all possible anchors, which is the same for all anchors from the same category, shown in~\cref{equ:postive_samples}.
\begin{equation}\label{equ:postive_samples}
    \mathbf{E_c^+} = \frac{1}{\left\lvert \mathcal{A}_c \right\rvert} \sum_{\textbf{E}_c \in \mathcal{A}_c}{\textbf{E}_c}.
\end{equation}
Following~\cite{wang2022semisupervised}, we similarly construct multiple negative samples $\textbf{E}_c^-$ for each anchor voxel.

Finally, for each anchor voxel containing one positive sample and $N-1$ negative samples, we propose the voxel-level InfoNCE loss~\cite{oord2019representation} (a variant of contrastive loss) $\mathcal{L}_C$ in~\cref{eq:contrastive_loss} to encourage maximal similarity between the anchor voxel and the positive sample, and the minimal similarity between the anchor voxel and multiple negative samples.
\begin{equation}
    \label{eq:contrastive_loss}
    \footnotesize
    \begin{split}
        \mathcal{L}_C &= -\frac{1}{C} \sum_{c=0}^{C-1} \underset{\mathbf{E}_c}{\mathbb{E}}\left[\log \frac{f\left(\textbf{e}_c, \textbf{e}_c^{+}, \tau\right)}{\sum_{\textbf{e}{_{c,j}^{-}} \in \textbf{E}{_c^{-}}} f\left(\textbf{e}_c, \textbf{e}_{c,j}^{-}, \tau\right)}\right] \\
        &= -\frac{1}{C} \sum_{c=0}^{C-1} \underset{\mathbf{E}_c}{\mathbb{E}} \left[ \log \frac{\exp\left(\left\langle \textbf{e}_c, \textbf{e}_c^+ \right\rangle / \tau \right)}{\exp\left(\left\langle \textbf{e}_c, \textbf{e}_c^+ \right\rangle / \tau \right) + \sum\limits_{j=1}^{N-1}{\exp \left(\left\langle \textbf{e}_{c}, \textbf{e}_{c,j}^- \right\rangle  / \tau \right)}} \right]
    \end{split}
\end{equation}
where $\left\langle \cdot, \cdot \right\rangle$ denotes cosine similarity. $\mathbf{e_c}$, $\mathbf{e_c^+}$ and $\textbf{e}_{c, j}^{-}$ denote the embedding, positive sample of the current anchor voxel, and embedding of the $j$-th negative sample of class $c$.

\subsection{Reflectivity-Based Test Time Augmentation}

\noindent To obtain minimal accuracy degradation despite very few weak labels, \eg, 1\% weakly-labeled ScribbleKITTI~\cite{Unal_2022_CVPR} dataset, we propose a Test Time Augmentation (TTA) that does not depend on any label, but only relies on a feature of original LiDAR points themselves. Also included in almost every LiDAR benchmark dataset for autonomous driving~\cite{Geiger2012,behley2019semantickittia,nuscenes2019,lyft2019,li2021durlara,Unal_2022_CVPR}, is the intensity of light reflected from the surface of an object at each point. In the presence of limited data labels in the semi-supervised learning case, this property of the material surface, normalized by distance to obtain the surface reflectivity in~\cref{equ:def_reflec}, could readily act as auxiliary information to identify different semantic classes. 

Our intuition is that reflectivity $R$, as a point-wise distance-normalized intensity feature, offers consistency across lighting conditions and range as:
\begin{equation}\label{equ:def_reflec}
    R = I r^{2} = \frac{S}{4 \pi r^2} \cdot r^{2} \propto S,
\end{equation}
where $S$ is the return strength of the LiDAR laser pulse, $I$ is the intensity and $r$ is the point distance from the source on the basis that scene objects with similar surface material, coating, and color characteristics will share similar $S$ returns.

\begin{figure}[thp]
    \centering
    \includegraphics[width=0.45\textwidth]{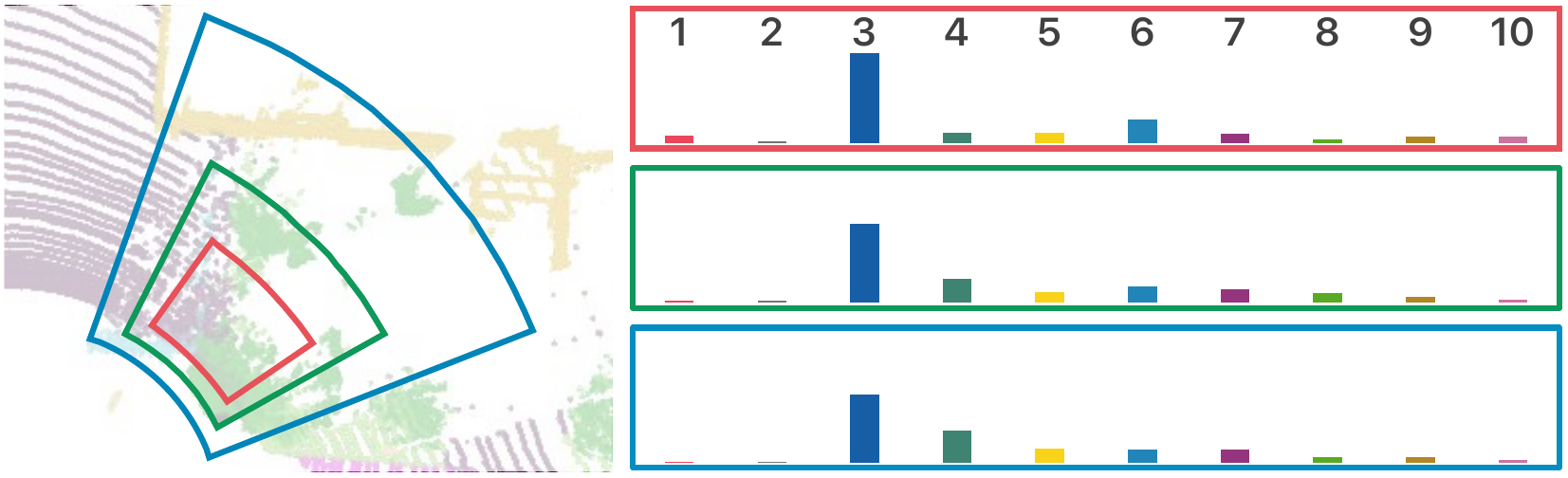}
    \caption{Coarse histograms of Reflec-TTA bins (not to scale). }
    \label{fig:reflec_tta}
    \vspace{-5pt}
\end{figure}

On this basis, we define our novel reflectivity-based Test Time Augmentation (Reflec-TTA) technique, as a substitute for label-dependent Pyramid Local Semantic-context (PLS) augmentation~\cite{Unal_2022_CVPR} during test-time as ground truth is not available. We append our point-wise reflectivity to the existing point features in order to enhance performance in presence of false or non-existent pseudo-labels at the distillation stage. As shown in~\cref{fig:reflec_tta},  and following~\cite{Unal_2022_CVPR}, we apply various sizes $s$ of bins in cylindrical coordinates to analyze the intrinsic point distribution of the LiDAR sensor at varying resolutions (shown in red, green and blue in ~\cref{fig:reflec_tta}). For each bin $b_i$, we compute a coarse histogram, $\mathbf{h}_i$:
\begin{equation}\label{equ:bins_histo}
    \begin{aligned}
        &\mathbf{h}_i& &\hspace{-10pt} = \left\{ h_i^{(k)} \mid k \in \left[1, N_{b}\right]  \right\} \in \mathbb{R}^{N_{b}}, \quad i \in \left[1, s\right], \\
        &h_i^{(k)}& &\hspace{-10pt} =\#\left\{R_j \in r_k, \; \forall j \mid p_j \in b_i\right\}, \\
        &r_k& &\hspace{-10pt} = [\; (k-1)/N_{b}, \; k/N_{b}\;), \quad k \in \left[\; 1, N_{b} \;\right].
    \end{aligned}
\end{equation}
The Reflec-TTA features $R^\circledast$ of all points $p_j \in b_j$ is further computed as the concatenation of the coarse histogram $\mathbf{h}_i$ of the normalized histogram:
\begin{equation}\label{equ:reflectta}
    R^\circledast=\left\{\mathbf{h}_i / \max\left(\mathbf{h}_i\right)  \mid i \in \left[1, s\right] \right\} \in \mathbb{R}^{s N_{b}}
\end{equation}
In the distillation stage, we append $R^\circledast$ to the input features and redefine the input LiDAR point cloud as the augmented set of points $P^\circledast = \left\{p \mid \left(x,y,z,I,R^\circledast\right) \in \mathbb{R}^{s N_{b}+4} \right\} $.

\subsection{\textbf{\textls[-2]{{\samplfull}}}}
\label{sec:MRFD}

\begin{figure}[htp]
    \hspace*{-12pt}
    \centering
    \includegraphics[width=0.52\textwidth]{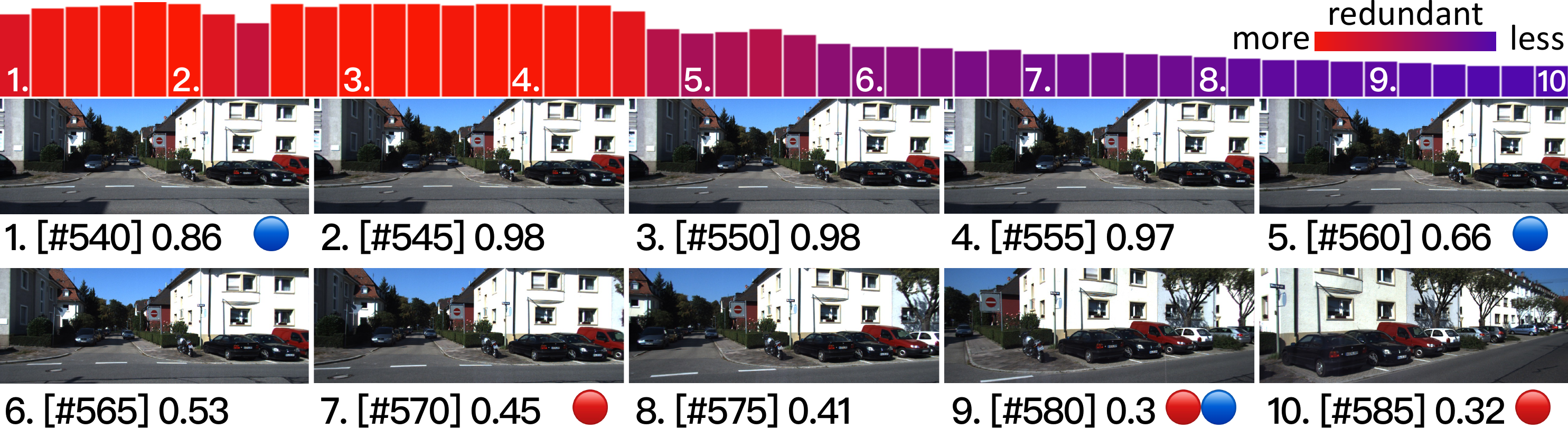}
    \caption{Illustration of LiDAR frame temporal correlation as \textit{[\#~frame~ID]~redundancy} with 5\% sampling on SemanticKITTI~\cite{behley2019semantickittia} ( sequence \texttt{00}) using uniform sampling (selected frames in \textcolor{fg5_blue}{\FilledSmallCircle}) and {\samplshort} strategy (\textcolor{fg5_red}{\FilledSmallCircle}).}
    \label{fig:mrfd_real}
    \vspace{-10pt}
\end{figure}

\noindent
Due to the spatio-temporal correlation of LiDAR point cloud sequences often captured from vehicles in %
metropolitan locales, many large-scale point cloud datasets demonstrate significant redundancy. Common datasets employ a frame rate of 10Hz~\cite{Geiger2012,nuscenes2019,behley2019semantickittia,lyft2019,Bijelic2020,sun2020scalability,li2021durlara}, and a number of concurrent laser channels (beams) of 32~\cite{nuscenes2019}, 64~\cite{Geiger2012,behley2019semantickittia,lyft2019,Bijelic2020,sun2020scalability} or 128~\cite{li2021durlara}. Faced with such large-scale, massively redundant training datasets, the popular practice of semi-supervised semantic segmentation approaches~\cite{tarvainen2017mean,french2020semisupervised,zou2018unsupervised,chen2021semisupervised,kong2022lasermix} is to uniformly sample 1\%, 10\%, 20\%, or 50\% of the available annotated training frames, without considering any redundancy attributable to temporary periods of stationary capture (\eg due to traffic, \cref{fig:mrfd_real}) or multi-pass repetition (\eg due to loop closure).

To extract a diverse set of frames, we propose a novel algorithm called {\samplfull} ({\samplshort}, \cref{alg:MRFD}) that determines spatio-temporal redundancy by analyzing the spatial-overlap within time-continuous LiDAR frame sequences. %
The key idea is that if spatial-overlap among some continuous frame sequence is high due to spatio-temporal redundancy, %
multiple representative frames can be sub-sampled for training, significantly reducing both training dataset size and redundant training computation. %

\begin{figure*}[htp]
    \centering
    \includegraphics[width=1\textwidth]{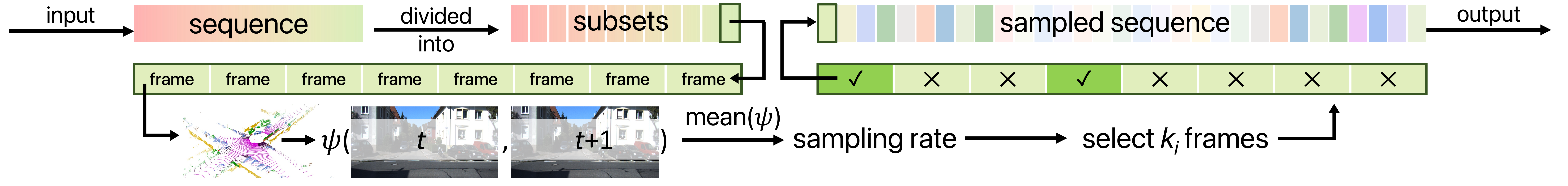}
    \caption{Overview of our proposed {\samplfull} approach.}
    \label{fig:mrfd_ex}
    \vspace{-8pt}
\end{figure*}

\cref{fig:mrfd_ex} shows an overview of {\samplshort}. It is conducted inside each temporal continuous LiDAR sequence $e$. First, we evenly divide $p$ point cloud frames in each sequence into $\left\lceil p/q \right\rceil$ subsets (containing $q$ frames). For each frame at time $t$ inside the subset, we find its corresponding RGB camera image in the dataset at time $t$ and $t+1$. To detect the spatio-temporal redundancy at time $t$, the similarity $\psi(t, t+1)$ between temporally adjacent frames are then computed via the Structural Similarity Index Measure (SSIM,~\cite{wang2004image}). We utilize the mean value of similarity scores between all adjacent frames in the current subset as a proxy to estimate the spatio-temporal redundancy present. A sampling rate is then determined according to this mean similarity for frame selection within this subset. This is repeated for all subsets in every sequence to construct our final set of sub-sampled LiDAR frames for training.

Concretely, as shown in~\cref{alg:MRFD}, we implement a {\samplshort} supervisor that determines the most informative assignments (\ie, the key point cloud frames) that the teacher and student networks should train on respectively. The {\samplshort} supervisor has an empirical supervisor function $\upsilon$, which decides the amount of assignments, \ie, the sampling rate corresponding to the extent of spatio-temporal redundancy.  
Using SSIM~\cite{wang2004image} as the redundancy function $\psi$ to measure the similarity between the RGB images associated with two adjacent point clouds, we define the empirical supervisor function $\upsilon$ with decay property $\upsilon (x) = \exp(-\beta x)$, where $\beta \in (0, +\infty)$ is the decay coefficient, and $x$ is the redundancy calculated from $\psi$. In this way, the higher the degree of spatio-temporal redundancy (as $\psi \rightarrow 1$), the lower the sampling rate our {\samplshort} supervisor will allocate, hence reducing the training set requirements for teacher and student alike.

\begin{algorithm}[htp]
    \footnotesize
    \SetKwInput{KwInput}{Input}
    \SetKwInput{KwOutput}{Return}
    \DontPrintSemicolon
    \SetAlgoLined
    \SetNoFillComment
    \caption{{\samplfull}.}
    \label{alg:MRFD}
    \SetAlgoVlined
    \KwInput{Point cloud frames pool $P$ (size of $p$), subset size $q$, redundancy function $\psi \in [0, 1]$ and empirical supervisor function~$\upsilon$.}
    Divide $P$ evenly into $\left\lceil p/q \right\rceil$ subsets $Q$. \\
    $D \gets$ empty dictionary. \\
    \ForAll{$e \gets 0:n_e-1$}{
        \tcp{loop for all sequences} 
        $C_e \gets \varnothing$ \tcp*{chosen point cloud frames} 
        \ForAll{$i \gets 0:\left\lceil p/q \right\rceil-1$}{
            \tcp{loop for subsets $Q$}
            $Q_{i, j} \gets$ $j$-th frame in subset $Q_i$. \\
            $\overline{M_i} \gets \frac{1}{q} \sum_{j=0}^{q-1}{\psi(Q_{i,j})}$. \tcp*{redundancy} 
            $k_i \gets \left\lceil \upsilon(\overline{M_i}) \cdot q\right\rceil $. \\
            $T_i \gets$ select $k_i$ frames in $Q_i$ with the smallest $M_i$.  \\
            $C_e \gets C_e \cup T_i$. 
        }
        Append key-value pair $(e, C_e)$ into $D$. 
    }
    \KwOutput{Dictionary $D$.}
\end{algorithm}

\begin{figure}[thp]
    \hspace{-0.48cm}
    \centering
    \includegraphics[width=0.5\textwidth]{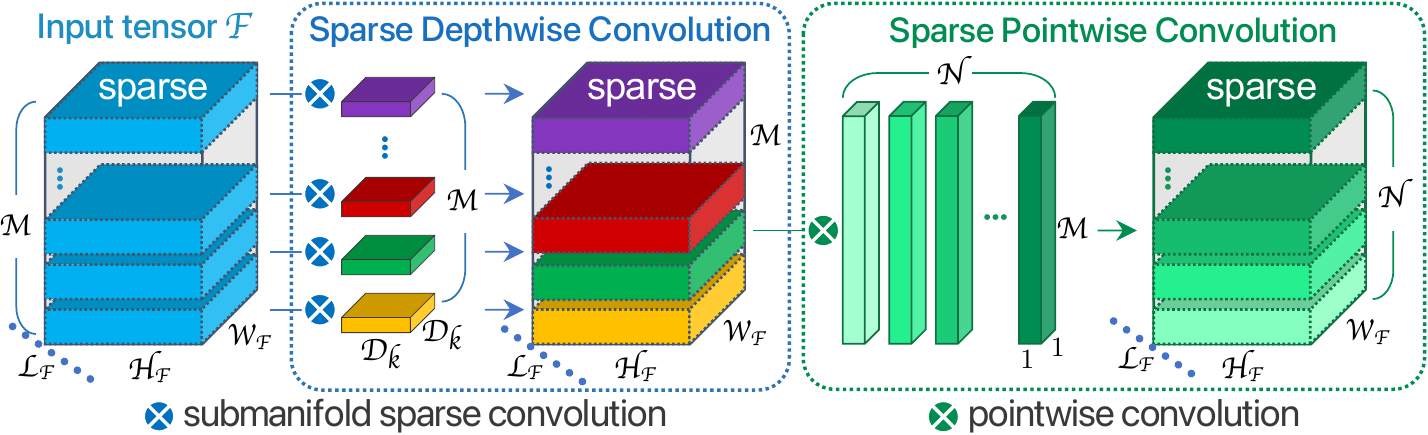}
    \caption{Illustration of the SDSC convolution module.}
    \label{fig:sdsc_module}
    \vspace{-0.5cm}
\end{figure}

\subsection{Sparse Depthwise Separable Convolution}
\label{sec:sdsc}

\noindent
Existing LiDAR point cloud semantic segmentation methods generally rely on a large-scale backbone architecture with tens of millions of trainable parameters ~\cite{hu2020randlanet,hou2022pointtovoxel,jaritz2021xmuda,Unal_2022_CVPR,yi2021complete,zhu2021cylindrical} due to the requirement for 3D (voxel-based) convolution operations, to operate on the voxelized topology of the otherwise unstructured LiDAR point cloud representation, which suffer from both high computational training demands and the risk of overfitting. Based on the observation that depthwise separable convolution has shown results comparable with regular convolution in tasks such as image classification but with significantly fewer trainable parameters~\cite{chollet2017xception,howard2017mobilenets,sandler2018mobilenetv2,tan2019efficientnet,masters2021making,howard2019searching}, here we pursue the use of such an approach within 3D point cloud semantic segmentation.

As such we propose the first formulation of sparse variant depthwise separable convolution~\cite{howard2017mobilenets} applied to 3D point clouds, namely Sparse Depthwise Separable Convolution (SDSC). SDSC combines the established computational advantages of sparse convolution for point cloud segmentation~\cite{graham20183d}, with the significant trainable parameter reduction offered by depthwise separable convolution~\cite{chollet2017xception}.

Our SDSC module, as outlined in~\cref{fig:sdsc_module}, initially takes a tensor $F \in \mathbb{R} ^{H_F \times W_F \times L_F \times M}$ as input, where $H_F$, $W_F$, $L_F$ and $M$ denote radius, azimuth, height in the cylinder coordinate~\cite{zhu2021cylindrical} and channels respectively. Firstly, a sparse depthwise convolution $\text{SDC}(M,M,D_k,s=1)$ is applied, with $M$ input and output feature planes, a kernel size of $D_k$ and stride $s$ in order to output a tensor $T \in \mathbb{R} ^{H_F \times W_F \times L_F \times M}$. Inside our sparse depthwise convolution, $M$ sparse spatial convolutions are performed independently over each input channel using submanifold sparse convolution~\cite{graham20183d} due to its tensor shape preserving property at no computational or memory overhead. Secondly, the sparse pointwise convolution $\text{SPC}(M,N,1,s=1)$ projects the channels output $T$ by the sparse depthwise convolution onto a new channel space, to mix the information across different channels. As a result, the sparse depthwise separable convolution $\text{SDSC}(M,N,D_k,s=1)$ is the compound of the sparse depthwise convolution and the sparse pointwise convolution, namely $\text{SDSC}(M,N,D_k,s=1)=\text{SDC}\circ \text{SPC}$.

Using a sparse voxelized input representation similar to~\cite{graham2015sparse}, and a series of such SDSC sub-modules we construct the popular Cylinder3D~\cite{zhu2021cylindrical} sub-architectures within our overall Mean Teacher architectural design (\cref{fig:model}). 

%% file: sections/4-exper.tex
\vspace{-0.3cm}   
\section{Evaluation}
\label{sec:experiments}
\noindent
We evaluate our proposed \textit{Less is More 3D} ({\ourmodel}) approach against %
state-of-the-art 3D point cloud semantic segmentation approaches using the SemanticKITTI~\cite{behley2019semantickittia} and ScribbleKITTI~\cite{Unal_2022_CVPR} benchmark datasets.

\input{tables/benchmark_2.tab}

\subsection{Experimental Setup}

\bdtitle{SemanticKITTI}~\cite{behley2019semantickittia} is a large-scale 3D point cloud dataset for semantic scene understanding with 20 semantic classes consisting of 22 sequences - [\texttt{00} to \texttt{10} as \textit{training}-split (of which \texttt{08} as \textit{validation}-split) + \texttt{11} to \texttt{21} as \textit{test}-split].

\bdtitle{ScribbleKITTI}~\cite{Unal_2022_CVPR} is the first scribble (\ie sparsely) annotated dataset for LiDAR semantic segmentation providing sparse annotations for the \textit{training} split of SemanticKITTI for 19 classes, with only 8.06\% of points from the full SemanticKITTI dataset annotated.

\bdtitle{Evaluation Protocol:} Following previous work~\cite{zhu2021cylindrical,jiang2021guided,Unal_2022_CVPR,kong2022lasermix}, we report performance on the SemanticKITTI and ScribbleKITTI {\trainset} for intermediate training steps, as this metric provides an indication of the pseudo-labeling quality, and on the {\validset} to assess the performance benefits of each individual component. Performance is reported using the mean Intersection over Union (mIoU, as \%) metric. For semi-supervised training, we report over both the benchmarks using the SemanticKITTI and ScribbleKITTI {\validset} under 5\%, 10\%, 20\%, and 40\% partitioning. We further report the relative performance of semi-supervised or scribble-supervised for ScribbleKITTI (SS) training to the fully supervised upper-bound (FS) in percentages (SS/FS) to further analyze semi-supervised performance and report the results for the fully-supervised training on both {\validset}s for reference. The trainable parameter count and number of multiply-adds (multi-adds) are additionally provided as a metric of computational cost.

\bdtitle{Implementation Details:} Training is performed using 4$\times$ NVIDIA A100 80GB GPU without pre-trained weights with a DDP shared training strategy~\cite{FairScale2021} to maintain GPU scaling efficiency, whilst reducing memory overhead significantly. Specific hyper-parameters are set as follows - Mean Teacher: $\kappa=0.99$; unreliable pseudo-labeling: $\lambda_C=0.3$, $\tau=0.5$; {\samplshort}: $\beta = \{7.45$, $5.72$, $4.00$, $2.28$, $0$\} for sampling \{5\%, 10\%, 20\%, 40\%, 100\%\} labeled training frames, assuming the remainder as unlabeled; Reflec-TTA: $N_b=10$, $s=3$ various Reflec-TTA bin sizes, following~\cite{Unal_2022_CVPR}, we set each bin $b_i = \left(\rho, \phi\right) \in \{ (20, 40), (40, 80), (80, 120)$\}.

\begin{figure}[thp]
    \centering
    \includegraphics[width=0.478\textwidth]{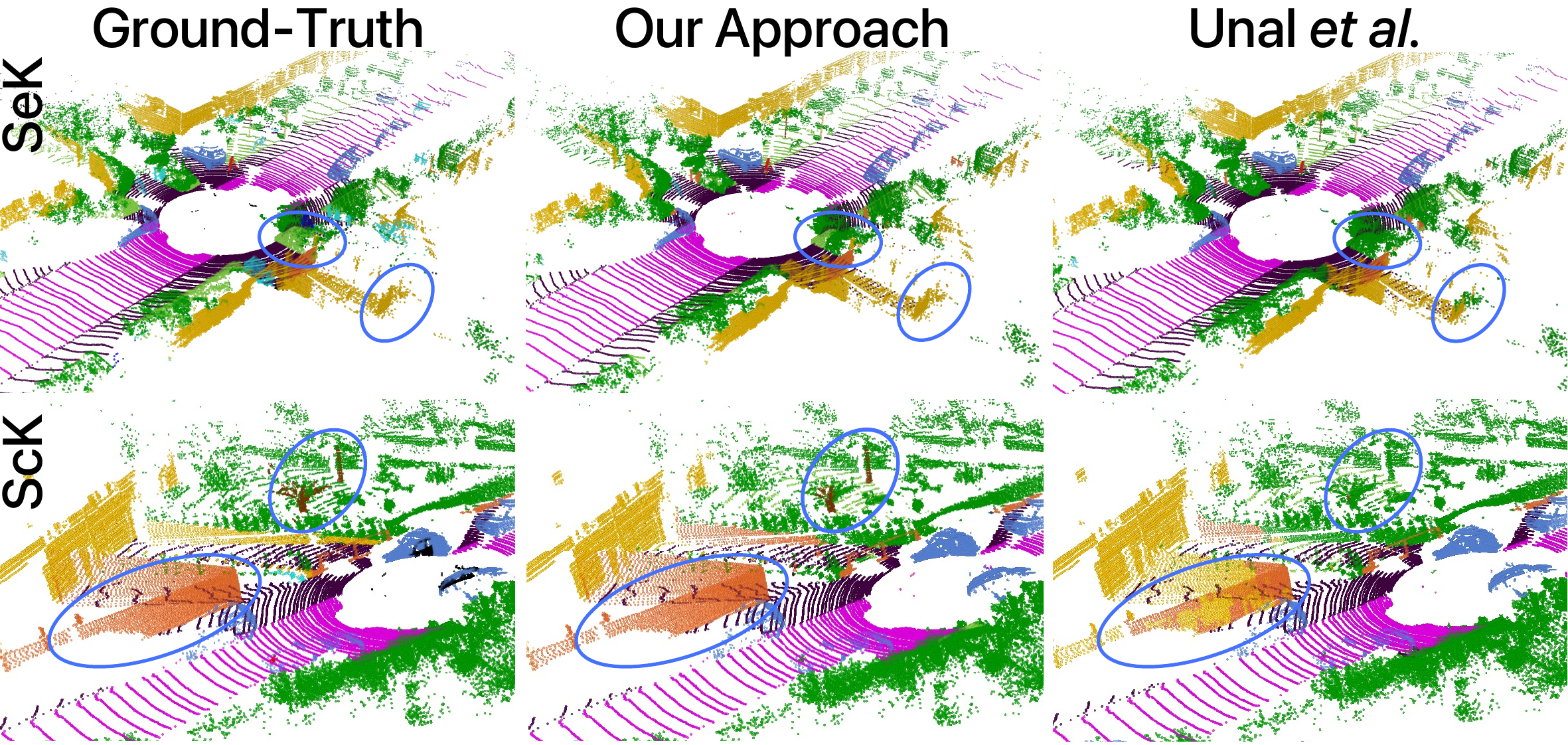}
    \caption{Comparing the 10\% sampling split of SemanticKITTI (SeK, first row) and ScribbleKITTI (ScK, second row) {\validset} with ground-truth (left), our approach (middle) and Unal \etal~\cite{Unal_2022_CVPR} (right) with areas of improvement highlighted.}
    \label{fig:visual_results}
    \vspace{-0.4cm}
\end{figure}

\input{tables/ablation.tab}

\vspace{-0.2cm}
\subsection{Experimental Results}
\vspace{-0.2cm}
\noindent
In~\cref{tab:benchmark_2}, we present the performance of our \textit{Less is More} 3D ({\ourmodel}) point cloud semantic segmentation approach both with ({\ourmodelsdsc}) and without ({\ourmodel}) SDSC in a side-by-side comparison with leading contemporary state-of-the-art approaches on the SemanticKITTI and ScribbleKITTI benchmark {\validset}s to illustrate our approach offers superior or comparable (within 1\% mIoU) performance across all sampling ratios. Furthermore, we present supporting qualitative results in~\cref{fig:visual_results}.

On SemanticKITTI, with a lack of available supervision, {\ourmodel} shows a relative performance (SS/FS) from $85.6\%$ ($5\%$-fully-supervised) to $91.1\%$ ($40\%$-fully-supervised), and {\ourmodelsdsc} from $85.3\%$ to $92.0\%$, compared to their respective fully supervised upper-bound. {\ourmodel}/{\ourmodelsdsc} performance is also less sensitive to reduced labeled data sampling compared with other methods.

Our model significantly outperforms on small ratio sampling splits, \eg, $5\%$ and $10\%$. {\ourmodel} shows up to $19.8\%$ and $18.9\%$ mIoU improvements whilst, with a smaller model size {\ourmodelsdsc} again shows significant mIoU improvements by up to $16.4\%$ and $15.5\%$ when compared with other range and voxel-based methods respectively.

\vspace{-0.15cm}
\subsection{Ablation Studies}
\vspace{-0.15cm}
\bdtitle{Effectiveness of Components.} In~\cref{tab:ablation} we ablate each component of {\ourmodel} step by step and report the performance on the SemanticKITTI {\trainset} at the end of training as an overall indicator of pseudo-labeling quality in addition to the corresponding {\validset}.

As shown in~\cref{tab:ablation}, adding unreliable pseudo-labeling (UP) in the distillation stage, we can increase the $valid$ mIoU by $+0.7\%$ on average in {\validset}. Appending reflectivity features (RF) in the training stage, we further improve the mIoU on the {\trainset} by $+0.7\%$ on average. Due to the improvements in training, the model generates a higher quality of pseudo-labels, which results to a $+0.5\%$ increase in mIoU in the {\validset}. If we disable reflectivity features in the training stage, applying Reflec-TTA in the distillation stage alone, we then get an average improvement of $+1.3\%$ compared with pseudo-labeling only. On the whole, enabling all reflectivity-based components (RF+RT) shows great improvements of up to $+2.8\%$ in $validation$ mIoU.
\input{tables/computation_cost.tab}
\input{tables/sampl.tab} 
\input{tables/pseudo.tab}
\input{tables/ref_vs_inten.tab}

Substituting the uniform sampling with our {\samplshort} strategy, we observe further average improvements of $+1.0\%$ and $+0.8\%$ on $training$ and $validation$ respectively (\cref{tab:ablation}).

Our SDSC module reduces the trainable parameters of our model by $57\%$, with a performance cost of $-0.7\%$ and $-1.4\%$ mIoU on $training$ and $validation$ respectively (\cref{tab:ablation}). Finally, we provide two models, one without SDSC ({\ourmodel}) and one with ({\ourmodelsdsc}), corresponding to the bottom two rows of~\cref{tab:ablation}.

\bdtitle{Effectiveness of SDSC module.} In ~\cref{tab:computation_cost}, we compare our {\ourmodel} and {\ourmodelsdsc} with recent state-of-the-art methods under 5\%-labeled semi-supervised training on the SemanticKITTI and ScribbleKITTI {\validset}s. {\ourmodelsdsc} outperforms the voxel-based methods~\cite{zhu2021cylindrical,Unal_2022_CVPR} with at least a \textbf{2.3}$\times$ reduction in model size. Similarly, with comparable model size~\cite{choy20194d,tang2020searching,yan20222dpass}, {\ourmodelsdsc} has higher mIoU in both datasets and up to \textbf{641}$\times$ fewer multiply-add operations.

\bdtitle{Effectiveness of {\samplshort} strategy.} In~\cref{tab:sampl}, we illustrate the effectiveness of our {\samplshort} strategy by comparing  {\ourmodel} with two widely-used strategies in semi-supervised training, \ie, random sampling and uniform sampling on SemanticKITTI~\cite{behley2019semantickittia} and ScribbleKITTI~\cite{Unal_2022_CVPR} {\validset}. Whilst uniform and random sampling have comparable results on both {\validset}s, simply applying our {\samplshort} strategy improves the baseline by $+0.90\%$, $+0.75\%$, $+0.60\%$ and $+0.55\%$ on SemanticKITTI under $5\%$, $10\%$, $20\%$ and $40\%$ sampling protocol respectively.
Furthermore, using corresponding range images of point cloud, rather than RGB images to compute the spatio-temporal redundancy within {\samplshort} (see {\samplshort}-R in~\cref{tab:sampl}), has no significant difference on the performance.

\bdtitle{Effectiveness of Unreliable Pseudo-Labeling.} In~\cref{tab:pseudo}, we evaluate selecting negative candidates with different reliability to illustrate the improvements of using unreliable pseudo-labels in semi-supervised semantic segmentation. The \textit{“Unreliable”} selecting of negative candidates outperforms other alternative methodologies, showing the positive performance impact of unreliable pseudo-labels.

\bdtitle{Effectiveness of Reflec-TTA.} In~\cref{tab:ablation}, we compare {\ourmodel} performance with and without Reflec-TTA and further experiment on the SemanticKITTI and ScribbleKITTI {\validset} in~\cref{tab:ref_vs_inten}. This demonstrates that the LiDAR point-wise intensity feature $I^\circledast$, in place of the distance-normalized reflectivity feature $R^\circledast$, offers inferior on-task performance. %

%% file: tables/benchmark_2.tab.tex
\begin{table*}[htp]
    \scriptsize
    \setlength{\abovecaptionskip}{0.05cm}
    \centering
    \caption{Comparative mIoU for Range- and Voxel-based methods using uniform sampling (U), sequential partition (P) and {\samplshort} sampling (S): \textbf{bold}/\underline{underlined} = \textbf{best}/\underline{2nd best}; $^*$ denotes reproduced result; -- denotes missing result due to unavailability from original authors.}
    \resizebox{\textwidth}{!}
    {
    \begin{tabular}{@{}C{0.5cm}|C{0.4cm}|R{1.55cm}L{1.05cm}|C{0.35cm}C{0.35cm}C{0.35cm}C{0.35cm}C{0.35cm}C{0.35cm}C{0.35cm}|C{0.35cm}C{0.35cm}C{0.35cm}C{0.35cm}C{0.35cm}C{0.35cm}C{0.35cm}}
    \topr
    \multirow{2}{*}{Repr.} & \multirow{2}{*}{Samp.} & \multirow{2}{*}{Method} &  & \multicolumn{7}{c|}
    {SemanticKITTI~\cite{behley2019semantickittia}} & \multicolumn{7}{c}{ScribbleKITTI~\cite{Unal_2022_CVPR}} \\
    &  &  &  & \hspace{0.5pt}1\% & \hspace{0.5pt}5\% & 10\% & 20\% & 40\% & 50\% & 100\% & \hspace{0.5pt}1\% & \hspace{0.5pt}5\% & 10\% & 20\% & 40\% & 50\% & 100\% \\
    \toprr

    \multirow{1}{*}{{Range}} & U & LaserMix~\cite{kong2022lasermix} & (2022) & 43.4 & -- & 58.8  & 59.4  &  \hspace{4.5pt}-- & 61.4 &\hspace{4.5pt}-- & 38.3 & \hspace{4.5pt}-- & 54.4  & 55.6 & \hspace{4.5pt}-- & 58.7 &\hspace{4.5pt}-- \\
    \midrule

    \multirow{5}{*}{{Voxel}} & U & Cylinder3D~\cite{zhu2021cylindrical} & (CVPR'21) & \hspace{4.5pt}-- & 45.4 & 56.1 & 57.8 & 58.7 & \hspace{4.5pt}-- & 67.8 & \hspace{4.5pt}-- & 39.2 & 48.0 & 52.1 & 53.8 & \hspace{4.5pt}-- &56.3 \\

    & U & LaserMix~\cite{kong2022lasermix} & (2022) & 50.6 & \hspace{4.5pt}-- & 60.0 & \underline{61.9} & \hspace{4.5pt}-- & 62.3 &\hspace{4.5pt}-- & 44.2 & \hspace{4.5pt}-- & 53.7 & 55.1 & \hspace{4.5pt}-- & 56.8 &\hspace{4.5pt}-- \\

    & P & Jiang \etal~\cite{jiang2021guided} & (ICCV'21)  & \hspace{4.5pt}-- & 41.8 & 49.9 & 58.8 & 59.9 & \hspace{4.5pt}-- & 65.8  &\hspace{4.5pt}--&\hspace{4.5pt}--&\hspace{4.5pt}--&\hspace{4.5pt}--&\hspace{4.5pt}-- &\hspace{4.5pt}--&\hspace{4.5pt}-- \\

    & U & Unal \etal~\cite{Unal_2022_CVPR} & (CVPR'22)  & \hspace{4.5pt}-- &49.9$^*$&58.7$^*$&59.1$^*$&60.9& \hspace{4.5pt}-- &\underline{68.2}$^*$& \hspace{4.5pt}-- &46.9$^*$&54.2$^*$&56.5$^*$&58.6$^*$& \hspace{4.5pt}-- &\underline{61.3} \\

    & S & {\ourmodelsdsc} & (ours) 
    & \underline{57.2}  %
    & \underline{57.6}  %
    & \underline{61.0}  %
    & 61.7  %
    & \underline{62.1}  %
    & \underline{62.7}  %
    & 67.5  %
    & \underline{55.8}  %
    & \underline{56.1}  %
    & \underline{56.9}  %
    & \underline{57.2}  %
    & \underline{58.9}  %
    & \underline{59.3}  %
    & 60.7  %
    \\

    & S & {\ourmodel} & (ours) 
    & \textbf{58.4}  %
    & \textbf{59.5}  %
    & \textbf{62.2}  %
    & \textbf{63.1}  %
    & \textbf{63.3}  %
    & \textbf{63.6}  %
    & \textbf{69.5}  %
    & \textbf{57.0}  %
    & \textbf{58.1}  %
    & \textbf{61.0}  %
    & \textbf{61.2}  %
    & \textbf{62.0}  %
    & \textbf{62.1}  %
    & \textbf{62.4}  %
    \\
    \bottomr
    \end{tabular}}
    \label{tab:benchmark_2}
\end{table*}

%% file: tables/ablation.tab.tex
\begin{table}[htp]
    \setlength{\abovecaptionskip}{0.05cm}
    \centering
    \caption{Component-wise ablation of {\ourmodel} (mIoU as \%, and \#parameters in millions, M) on SemanticKITTI~\cite{behley2019semantickittia} \textit{training} and \textit{validation} sets where UP, RF, RT, ST, SD denote Unreliable Pseudo-labeling, Reflectivity Feature, Reflec-TTA, {\samplshort}, and SDSC module respectively.}
    \resizebox{0.48\textwidth}{!}{
        \begin{tabular}{@{}C{0.25cm}C{0.25cm}C{0.25cm}C{0.25cm}C{0.25cm}|C{0.45cm}C{0.45cm}C{0.45cm}C{0.45cm}|C{0.45cm}C{0.45cm}C{0.45cm}C{0.45cm}|c@{}}
            \topr 
            \multirow{2}{*}{UP} & \multirow{2}{*}{RF} & \multirow{2}{*}{RT} & \multirow{2}{*}{ST} & \multirow{2}{*}{SD} & \multicolumn{4}{c|}{Training mIoU (\%)} & \multicolumn{4}{c|}{Validation mIoU (\%)} & {\#Params} \\
            &   &   &   &   & 5\% & 10\% & 20\% & 40\% & 5\% & 10\% & 20\% & 40\% & (M) \\
    \midr

    &  &  &  &  
    & 82.8    %
    & 87.5    %
    & 87.8    %
    & 88.2    %
    & 54.8    %
    & 58.1    %
    & 59.3    %
    & 60.8    %
    & 49.6   \\

    \checkmark &  &  &  &  & -- & -- & -- & -- 
    & 55.9    %
    & 58.8    %
    & 59.9    %
    & 61.2    %
    & 49.6   \\
    
    \midr
    \checkmark & \checkmark &  &  & 
    & 83.6    %
    & 88.3    %
    & 88.7    %
    & 89.1    %
    & 56.8    %
    & 59.6    %
    & 60.5    %
    & 61.4    %
    & 49.6   \\

    \checkmark &  & \checkmark &  &  & -- & -- & -- & -- 
    & 57.5    %
    & 59.8    %
    & 61.2    %
    & 62.6    %
    & 49.6   \\
    
    \checkmark & \checkmark & \checkmark &  &  & -- & -- & -- & -- 
    & 58.7    %
    & 61.3    %
    & 62.4    %
    & 62.8    %
    & 49.6   \\

    \midrule
    \checkmark & \checkmark & \checkmark & \checkmark & 
    & \textbf{85.2}    %
    & \textbf{89.1}    %
    & \textbf{89.5}    %
    & \textbf{89.7}    %
    & \textbf{59.5}    %
    & \textbf{62.2}    %
    & \textbf{63.1}    %
    & \textbf{63.3}    %
    & 49.6   \\

    \checkmark & \checkmark & \checkmark & \checkmark & \checkmark
    & 83.8    %
    & 88.6    %
    & 89.0    %
    & 89.2    %
    & 57.6    %
    & 61.0    %
    & 61.7    %
    & 62.1    %
    & \textbf{21.5}   \\
    \bottomr
    \end{tabular}}
    \label{tab:ablation}
    \vspace{-10pt}
\end{table}

%% file: tables/computation_cost.tab.tex
\begin{table}[htp]
    \scriptsize
    \vspace{-3pt}
    \setlength{\abovecaptionskip}{0.05cm}
    \centering
    \caption{The computation cost and mIoU (in percentage) under 5\%-labeled training results on SemanticKITTI (SeK) and ScribbleKITTI (ScK) {\validset}.}
{\begin{tabular}{lcrcc}
\toprule
Method & \# Parameters & \hspace{-6pt} \# Mult-Adds & SeK~\cite{behley2019semantickittia} & ScK~\cite{Unal_2022_CVPR} \\
\midrule
Cylider3D~\cite{zhu2021cylindrical} & 56.3 & 476.9M & 45.4 & 39.2 \\
Unal~\etal~\cite{Unal_2022_CVPR} & 49.6 & 420.2M & 49.9 & 46.9 \\
2DPASS~\cite{yan20222dpass} & 26.5 & \underline{217.4M} & 51.7 & 45.1 \\
MinkowskiNet~\cite{choy20194d} & 21.7 & 114.0G & 42.4 & 35.8 \\
SPVNAS~\cite{tang2020searching} & \textbf{12.5} & 73.8G & 45.1 & 38.9 \\
{\ourmodelsdsc} (ours) & \underline{21.5} & \textbf{182.0M} & \underline{57.6} & \underline{54.7} \\
{\ourmodel} (ours) & 49.6 & 420.2M & \textbf{59.5} & \textbf{58.1} \\
\bottomrule
\end{tabular}}
\label{tab:computation_cost}
\vspace{-10pt}
\end{table}

%% file: tables/sampl.tab.tex
\begin{table}[htp]
    \scriptsize
    \vspace{-3pt}
    \centering
    \setlength{\abovecaptionskip}{0.05cm}
    \caption{Effects of {\samplshort} sampling on SemanticKITTI and ScribbleKITTI {\validset} (mIoU as \%).}
{\begin{tabular}{c|cccc|cccc}
\toprule
\multirow{2}{*}{Sampling} & \multicolumn{4}{c|}{SemanticKITTI~\cite{behley2019semantickittia}} & \multicolumn{4}{c}{ScribbleKITTI~\cite{Unal_2022_CVPR}} \\
& 5\% & 10\% & 20\% & 40\% & 5\% & 10\% & 20\% & 40\% \\
\midrule 

Random 
& 58.5   %
& 61.6   %
& 62.6   %
& 62.7   %
& 57.1   %
& 60.3   %
& \underline{60.5}   %
& 60.9\\ %

Uniform  
& 58.7   %
& 61.3   %
& 62.4   %
& 62.8   %
& 56.9   %
& 60.6   %
& 60.3   %
& 61.0 \\

{\samplshort}-R
& \underline{59.1}                %
& \textbf{62.4}       %
& \underline{62.9}                %
& \textbf{63.4}       %
& \underline{58.0}                %
& \underline{60.7}                %
& \textbf{61.2}       %
& \underline{61.8} \\             %

{\samplshort}  & \textbf{59.5}  & \underline{62.2} & \textbf{63.1} & \underline{63.3} & \textbf{58.1} & \textbf{61.0}  & \textbf{61.2} & \textbf{62.0} \\

\bottomrule
\end{tabular}}
\vspace{-10pt}
\label{tab:sampl}
\end{table}

%% file: tables/pseudo.tab.tex
\begin{table}[H]
    \scriptsize
    \vspace{-3pt}
    \setlength{\abovecaptionskip}{0.05cm}
    \centering
\caption{Effects of differing reliability using pseudo voxels on SemanticKITTI {\validset}, measured by the entropy of voxel-wise prediction. %
\textit{Unreliable} and \textit{Reliable}: selecting negative candidates %
with top $20 \%$ highest entropy scores and bottom $20 \%$ counterpart respectively. \textit{Random}: sampling randomly regardless of entropy.
}
{
\begin{tabular}{c|cc|cc|cc}
\toprule 
 \multirow{2}{*}{Ratio} & \multicolumn{2}{c|}{Unreliable} & \multicolumn{2}{c|}{Reliable} & \multicolumn{2}{c}{Random} \\
 & mIoU & SS/FF & mIoU & SS/FF & mIoU & SS/FF \\
\midrule 
 5\% & \textbf{59.5} & \textbf{85.6} & 57.2 & 82.3 & 56.4 & 81.2 \\
10\% & \textbf{62.2} & \textbf{89.5} & 60.8 & 87.5 & 59.7 & 85.9 \\
20\% & \textbf{63.1} & \textbf{90.8} & 61.4 & 88.3 & 60.5 & 87.1 \\
40\% & \textbf{63.3} & \textbf{91.1} & 62.8 & 90.4 & 61.3 & 88.2 \\
 \bottomrule
\end{tabular}
\label{tab:pseudo}
}
\vspace{-10pt}
\end{table}

%% file: tables/ref_vs_inten.tab.tex
\begin{table}[!h]
    \scriptsize
    \vspace{-3pt}
    \centering
    \setlength{\abovecaptionskip}{0.05cm}
    \caption{Reflectivity (Reflec-TTA) vs. Intensity (intensity-based TTA) on \textls[-45]{SemanticKITTI and ScribbleKITTI} {\validset} (mIoU, \%).}
{\begin{tabular}{c|cccc|cccc}
\toprule
\multirow{2}{*}{TTA} & \multicolumn{4}{c|}{SemanticKITTI~\cite{behley2019semantickittia}} & \multicolumn{4}{c}{ScribbleKITTI~\cite{Unal_2022_CVPR}} \\
& 5\% & 10\% & 20\% & 40\% & 5\% & 10\% & 20\% & 40\% \\
\midrule 
Intensity 
& 56.2   %
& 59.1   %
& 59.8   %
& 60.9   %
& 55.7   %
& 57.5   %
& 57.9   %
& 59.2 \\
Reflectivity  & \textbf{59.5}  & \textbf{62.2} & \textbf{63.1} & \textbf{63.3} & \textbf{58.1} & \textbf{61.0}  & \textbf{61.2} & \textbf{62.0} \\
\bottomrule
\end{tabular}}
\label{tab:ref_vs_inten}
\vspace{-10pt}
\end{table}

%% file: sections/5-concl.tex
\vspace{-0.2cm}   
\section{Conclusion}
\vspace{-0.2cm}  
\label{sec:conclusion}

\noindent
This paper presents an efficient semi-supervised architecture for 3D point cloud semantic segmentation, which achieves \textit{more} in terms of performance with \textit{less} computational costs, \textit{less} annotations, and \textit{less} trainable model parameters (\ie, \textit{Less is More}, LiM3D). Our architecture consists of three novel contributions: the SDSC convolution module,  the {\samplshort} sampling strategy, and the pseudo-labeling method informed by LiDAR reflectivity. These individual components can be applied to any 3D semantic segmentation architecture to reduce the gap between semi or weakly-supervised and fully-supervised learning on task performance, whilst managing model complexity and computation costs.